# Confidence-Aware SSD Ensemble with Weighted Boxes Fusion for Weapon Detection

Atharva Jadhav
*Information Technology*
Thadomal Shahani Engineering College
Mumbai, India
me.atharvajadhav@gmail.com

Arush Karekar
*Information Technology*
Thadomal Shahani Engineering College
Mumbai, India
arushhk@gmail.com

Manas Divekar
*Information Technology*
Thadomal Shahani Engineering College
Mumbai, India
manas.divekar76@gmail.com

Shachi Natu
*Information Technology*
Thadomal Shahani Engineering College
Mumbai, India
shachi.natu@thadomal.org

*Abstract*— The safety and security of public spaces is of vital importance, driving the need for sophisticated surveillance systems capable of accurately detecting weapons, which are often hampered by issues like partial occlusion, varying lighting, and cluttered backgrounds. While single-model detectors are advanced, they often lack robustness in these challenging conditions. This paper presents the hypothesis that ensemble of Single Shot Multibox Detector (SSD) models with diverse feature extraction backbones can significantly enhance detection robustness. To leverage diverse feature representations, individual SSD models were trained using a selection of backbone networks: VGG16, ResNet50, EfficientNet, and MobileNetV3. The study is conducted on a dataset consisting of images of three distinct weapon classes: guns, heavy weapons and knives. The predictions from these models are combined using the Weighted Boxes Fusion (WBF) method, an ensemble technique designed to optimize bounding box accuracy. Our key finding is that the fusion strategy is as critical as the ensemble's diversity, a WBF approach using a 'max' confidence scoring strategy achieved a mean Average Precision (mAP) of 0.838. This represents a 2.948% relative improvement over the best-performing single model and consistently outperforms other fusion heuristics. This research offers a robust approach to enhancing real-time weapon detection capabilities in surveillance applications by demonstrating that confidence-aware fusion is a key mechanism for improving accuracy metrics of ensembles.

## I. INTRODUCTION

The sanctity of public spaces is critical to the well-being and stability of any society. These spaces are essential for social and economic activity, yet their vulnerability to violence, particularly with the proliferation of firearms, necessitates robust security measures. Ensuring safety in these environments is paramount for societal well-being and stability. Therefore, the development of proactive surveillance systems to detect and mitigate potential threats is crucial.

Traditional public surveillance, typically consisting of human-operated CCTV systems, helps in post-incident apprehension but lacks real-time prevention capabilities, leading to delays and potential escapes [1]. These methods also suffer from the need for constant human presence, and resulting human error and bias [2]. Automated systems, leveraging advanced algorithms and machine learning, offer real-time detection, continuous monitoring, and objective analysis, enabling proactive intervention and overcoming the limitations of human-based surveillance [3]. Their benefits have also been observed in several other domains, such as disease surveillance [4].

While the recent progress towards foundational Multimodal Large Language Models (MLLMs) excels at high-level visual reasoning, their performance in object detection tasks still lags behind classical computer vision systems [5,23,24,26]. Computer vision-based object detection systems, particularly those utilizing deep learning, have shown promise in various surveillance applications. The Single Shot MultiBox Detector (SSD) [6] is an object detection algorithm that balances the speed of single-stage detectors with the accuracy of two-stage methods. SSD predicts bounding boxes and class probabilities using feature maps from a convolutional neural network. It detects objects at various scales and aspect ratios by employing anchor boxes to improve localization accuracy, predicting offsets rather than direct coordinates. SSD is used in applications like object detection [7], traffic density estimation [8], and waste detection [9], often with backbones such as ResNet and MobileNet.

We present an ensemble strategy based on Weighted Boxes Fusion (WBF) applied to the SSD framework for weapon detection. SSD models were trained with diverse backbones, chosen for their complementary strengths. Individual model performance was optimized by experimenting with Non-Maximum Suppression (NMS) thresholds. Model weights within the WBF process were meticulously tuned to improve the weapon detection system's accuracy and robustness, particularly in conditions with poor image quality or partial concealment. Further sections of this paper will explore the relevant prior research in Section 2, followed by a detailed description of the dataset used for training and evaluation in Section 3, a comprehensive explanation of our proposed methodology in Section 4, the presentation and analysis of our experimental results in Section 5, and finally, the concluding remarks and future scope in Section 6.

## II. RELATED WORK

The use of model ensembles has proven effective to enhance object detection performance. Roman Solovyev, Weimin Wang and Tatiana Gabruseva [10] proposed Weighted Boxes Fusion (WBF), which uses confidence scores to average bounding boxes. Demonstrated on Open Images and COCO, WBF improved localization accuracy and showed strong performance in detection challenges. It is most effective with accurate models but may perform worse than NMS when many overlapping boxes have varying confidence scores.

In a related effort, Angela Casado-Garcia and Jonathan Heras [11] introduced a novel ensemble algorithm applicable to various object detection models. They reported performance improvements of up to 10% over baseline models, providing a versatile tool for boosting object detection performance. However, this study solely focuses on the NMS ensemble technique and does not discuss how other techniques like WBF, soft NMS or NMW could affect results.

Several works have focused on improving the SSD architecture itself. Xinbiao Gao, Junhua Xu, Chuan Luo, Jun Zhou, Panling Huang and Jianxin Deng [12] proposed R-SSD, an improved SSD algorithm. The authors replaced VGG16 with ResNet50 for better feature extraction, increased input resolution, and optimized anchor box aspect ratios. This resulted in a 7% improvement over standard SSD, and high detection confidence for practical applications. However, R-SSD's increased input resolution and deeper backbone (ResNet50) likely lead to higher computational cost and potentially slower inference speed, posing challenges for real-time deployment.

Dalmar Dakari Aboyomi, and Cleo Daniel [13] discussed a comprehensive comparative analysis of three widely used object detection algorithms: YOLO, SSD, and Faster R-CNN. This study evaluated these methods on a standardized dataset, focusing on detection accuracy, processing speed, and computational resource demands. The findings revealed YOLO's strength in real-time applications, SSD's balanced performance, and Faster R-CNN's superior accuracy. This analysis gave valuable insights for selection of an appropriate object detection algorithm based on specific application requirements.

In response to the growing need for real-time weapon detection in public spaces, Ayush Thakur, Akshat Shrivastav, Rohan Sharma, Triyank Kumar, and Kabir Puri [14] developed a YOLOv8 based system capable of identifying firearms and edged weapons in live video feeds. The model achieved strong results in terms of precision, recall, F1-score, and mean Average Precision (mAP), demonstrating its ability to reliably distinguish weapons. Their current research is limited by the dataset's scope in terms of environmental diversity and weapon variations, and further improvements could be made in inference speed and the integration of multi-modal data.

Haribharathi Sivakumar, Vijay Arvind.R, Pawan Ragavendhar V. and G. Balamurugan [15] proposed a real-time weapon detection pipeline using an ensemble of CNNs with diverse architectures, trained on minimally overlapping mini-batches. Their approach achieved a 5% improvement in accuracy, specificity, and recall over existing weapon detection systems, demonstrating the effectiveness of their ensemble strategy.

To enhance object detection accuracy in two-stage detectors like Faster R-CNN, Jinsu Lee, Sang-Kwang Lee and Seong-Il Yang [16] proposed an ensemble method with novel model selection and box voting strategies. They selected models based on overall mAP and per-class AP, and refined confidence scores using weighted box voting based on class-specific AP. Tested on the PASCAL VOC 2012 dataset, their approach significantly improved mAP over traditional ensembles.

Despite these advances, few studies have explored ensemble SSD architectures specifically optimized for weapon detection, especially with confidence-aware fusion strategies like WBF. This paper aims to bridge this gap by proposing an SSD-based ensemble framework with multiple CNN backbones, fused using WBF, to improve detection robustness in diverse environments.

### III. DATASET

This study utilizes a curated dataset specifically constructed for weapon detection by merging two publicly available datasets from Roboflow [17] [18]. The final dataset comprises 15,251 images spanning three distinct weapon classes: guns, heavy weapons (e.g., rifles, shotguns), and knives.

It encompasses a diverse range of images capturing weapons under varying conditions, including different lighting, backgrounds, and object orientations. This variability is crucial for training a model capable of generalizing to real-world scenarios. To maintain compatibility with established object detection benchmarks, the dataset's annotation structure mirrors the widely used Pascal VOC 2007 format, including class-specific bounding boxes for all three weapon categories. The dataset is partitioned into training (≈80%), validation (≈10%) and testing (≈10%) subsets to ensure robust model training and unbiased performance evaluation. The distribution of weapon classes is as follows:

- Training Set: Gun: 4063 images, Heavy Weapon: 4009 images, Knife: 4111 images
- Validation Set: Gun: 494 images, Heavy Weapon: 543 images, Knife: 484 images
- Testing Set: Gun: 523 images, Heavy Weapon: 549 images, Knife: 475 images
- Total images: Gun: 5080, Heavy Weapon: 5101, Knife: 5070

The dataset maintains a relatively balanced class distribution, with minor variations reflecting real-world data patterns.

### IV. METHODOLOGY

In this study, we propose an ensemble-based approach to enhance weapon detection by integrating multiple SSD models, each utilizing a distinct backbone network. This strategy leverages the diverse feature extraction capabilities of various backbones to improve detection accuracy and robustness. To combine the outputs of these models, we employ the Weighted Boxes Fusion (WBF) method, which refines bounding box predictions by merging overlapping boxes rather than suppressing them, thereby retaining more information and improving localization precision.

Single-shot object detection has gained popularity due to the speed limitations of traditional two-stage frameworks. It processes the entire image in a single pass, improving computational efficiency by eliminating the need for objectness based pruning and applying object classifiers and regressors densely. This approach directly predicts bounding boxes and class probabilities from feature maps extracted by convolutional backbones.

This study utilizes several CNN backbones for SSD: VGG16 [19], ResNet50 [20], MobileNet [21], and EfficientNet [22], each offering distinct strengths. VGG16 provides a deep architecture with small convolutional filters for rich feature extraction, aiding in precise object localization. ResNet50 employs identity shortcut connections to improve information flow and enhance optimization, enabling effective feature extraction for object localization and classification. MobileNet uses depth-wise separable convolutions for a lightweight and efficient architecture, suitable for resource-constrained devices and real-time applications, with a potential trade-off in accuracy. EfficientNet offers a balance of accuracy and efficiency through compound scaling of network dimensions, optimizing feature extraction and enabling competitive accuracy with faster inference speeds.

We utilize the SSD backbones to develop our ensemble approach using the WBF method, following the methodology in [10]. Unlike traditional Non-Maximum Suppression (NMS), which discards overlapping boxes, WBF fuses them based on their confidence scores, producing more accurate bounding boxes. It clusters predictions by Intersection Over Union (IoU) and computes weighted averages for coordinates and gives greater influence to higher confidence predictions. WBF also mitigates errors from slightly misaligned boxes by merging them into a more accurate result, with the final confidence score adjusted by the number of contributing boxes and models.

A key methodological decision in this work was the adoption of SSD as the common detection architecture integrated with diverse backbones. While optimized pairings like EfficientNet-EfficientDet exist, our core hypothesis focuses on the benefits of fusing features from varied backbones. Using a consistent SSD head isolates the impact of each backbone's feature extraction capabilities without interference from differing detection mechanisms. This uniformity ensures that the WBF ensemble operates on comparable outputs, effectively leveraging the complementary strengths of the backbones.

Fig. 1 shows the process of obtaining the detection outputs using our ensemble. Here is a detailed description of the process:

Step 1: The process starts with an input image consisting of the weapon which is provided to the model.

Step 2: The input image is fed into four different object detection models which consist of VGG16, ResNet50, MobileNet and EfficientNet as backbones. Each model is based on SSD architecture but uses a different backbone for detecting weapons in the input image.

Step 3: Each model produces a set of detections which contains:

- cls (class label): This indicates the type of weapon which is detected. Class 0 represents gun, Class 1 represents heavy weapon, Class 2 represents knife.
- cfs (confidence score): This indicates the certainty of

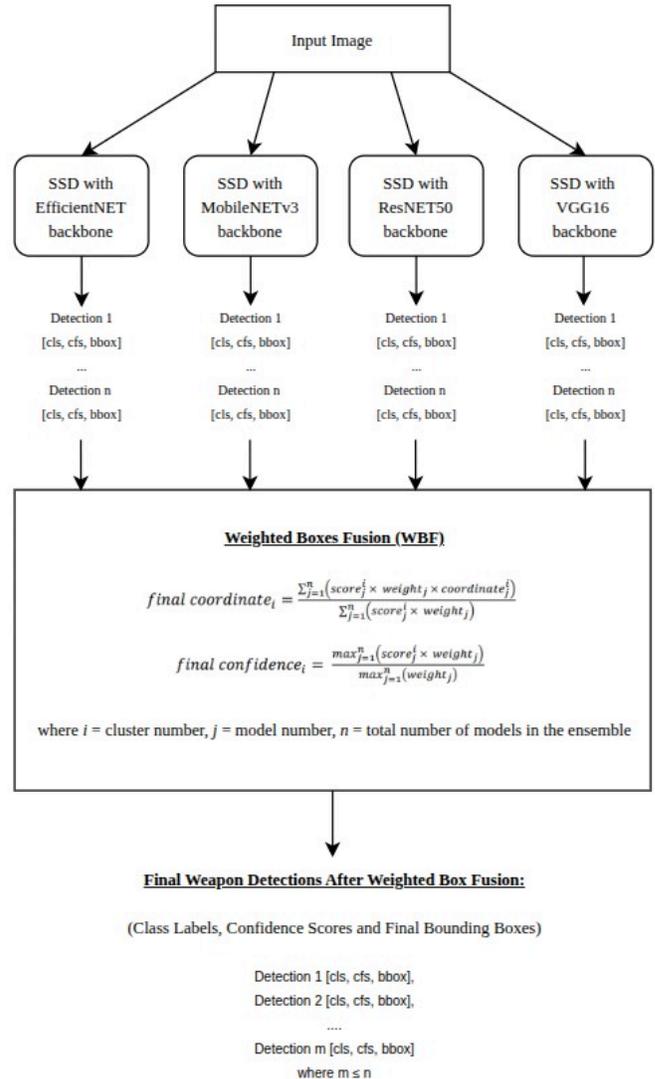

Fig. 1. Proposed Multi-Backbone SSD Ensemble with WBF

model that the detected object is a weapon.

- bbox (bounding box): Represented by coordinates of a box surrounding the detected weapon in the image

Each model detects multiple bounding boxes along with their corresponding class labels and confidence scores for that image.

Step 4: Weighted Boxes Fusion: In this step, the model clusters the output bounding boxes of the individual detectors based on a tuned IoU threshold and then combines the bounding box and confidence predictions to generate a more accurate and reliable final detection.

Weighted Averaging of Coordinates: For detected object in the $i^{th}$ cluster, each bounding box coordinate from all the models are combined using a weighted average. This means that detections with higher confidence scores contribute more to the final bounding box and confidence score. The weights are predetermined by the mAP of each model.

$$final\ coordinate_i = \frac{\sum_{j=1}^{n}(score_j^i \times weight_j \times coordinate_j^i)}{\sum_{j=1}^{n}(score_j^i \times weight_j)} \quad (1)$$

- $final\ coordinate_i$ is the final bounding box coordinate $(x_1^i, y_1^i, x_2^i, y_2^i)$ of the $i^{th}$ detection cluster.
- $score_j^i$ is the confidence score of the detected weapon in the $i^{th}$ detection cluster from the $j^{th}$ model.
- $weight_j$ is a predefined weight assigned to the $j^{th}$ model.
- $coordinate_j^i$ is the bounding box coordinate $(x_1, y_1, x_2, y_2)$ of the detected weapon in the $i^{th}$ detection cluster from the $j^{th}$ model.
- n is the total number of models used in the ensemble.

Confidence Calculation: The final confidence score for each fused detection is calculated based on the maximum weighted confidence among the individual detections.

$$final\ confidence_i = \frac{max_{j=1}^{n}(score_j^i \times weight_j)}{max_{j=1}^{n}(weight_j)} \quad (2)$$

- $final\ confidence_i$ is the final confidence score of the $i^{th}$ detection cluster

Step 5: Final Weapon Detection After WBF: The output of WBF is a refined set of weapon detection where each detection is represented by [cls, cfs, bbox].

- cls: This represents the class label indicating the type of weapon detected.
- cfs: Represents the final confidence score after the ensemble method.
- bbox: Represents the final bounding box coordinates after ensembling.

The number of final detections (m) may be less than or equal to the total number of detections from any of the individual models (n), as the fusion process may eliminate redundant or low-confidence detections.

Fig. 2 illustrates the performance of our proposed ensemble model in comparison to its constituent, individual models on a standard image.

## V. RESULTS

In this section, we look at how each of our SSD models perform individually and how they work together when combined using WBF. We use mAP to measure accuracy, and then perform a detailed comparison of weapon detection with various backbones, fusion strategies, and confidence scoring methods.

### A. Individual Model Performance

We first evaluated the performance of single-shot detectors based on the SSD framework using the four CNN backbones as described above, shown in Table 1. For individual models, the Non-Maximum Suppression (NMS) threshold was varied from 0.35 to 0.75, and the individual model mAP's were recorded for each backbone. We can observe from Table 1 that increasing the NMS threshold generally improves mAP up to a certain point but beyond that point, the improvement stops and, in some cases,

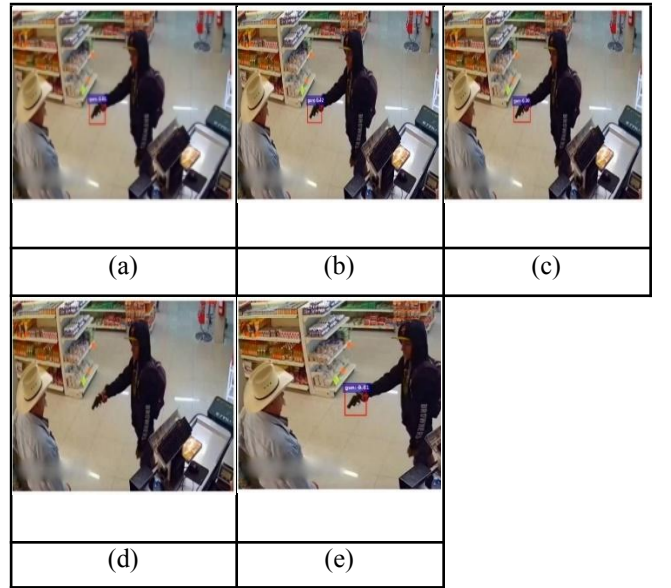

Fig. 2. (a) SSD with VGG16, confidence score of 0.81 (b) SSD with ResNet50, confidence score of 0.42 (c) SSD with EfficientNet, confidence score of 0.30 (d) SSD with MobileNet, no weapon detected (e) Our proposed Ensemble model, confidence score of 0.81

the mAP even reduces. VGG16 gave the best accuracy (mAP) compared to the other models at all the NMS settings tested. The similar results on both training and testing data show that the models learned well and didn't overfit.

Based on these experiments, an NMS threshold of 0.55 was selected for the final individual models, as it yielded the highest average mAP across all backbone networks. This selection was made through a thorough grid search to optimize the overall detection accuracy of the weapon detection system.

### B. Ensemble Performance with Weighted Boxes Fusion

To test how well the ensemble approach worked, we combined the predictions from the individual SSD models using the WBF method. Instead of just removing overlapping boxes, WBF merges them based on confidence scores. For this, we used a fixed IoU threshold for WBF of 0.5 to group the overlapping boxes before combining them. This value was chosen after a thorough grid search, during which we also tried out the following weighting strategies for WBF to fine-tune performance:

Table 1. Individual Model mAP at Selected NMS Thresholds. The results indicate that increasing the NMS threshold generally enhances the detection quality, with VGG16 consistently achieving the highest mAP among the individual backbones.

| NMS Threshold | mAP Scores | | | |
|---|---|---|---|---|
| | VGG16 | ResNet50 | EfficientNet | MobileNet |
| 0.35 | 0.7953 | 0.7222 | 0.5821 | 0.6837 |
| 0.4 | 0.7991 | 0.7197 | 0.6096 | 0.6754 |
| 0.45 | 0.799 | 0.7252 | 0.5875 | 0.6776 |
| 0.5 | 0.8004 | 0.7166 | 0.5838 | 0.6649 |
| 0.55 | 0.8137 | 0.7428 | 0.6028 | 0.6656 |
| 0.65 | 0.8049 | 0.7443 | 0.5829 | 0.6598 |
| 0.7 | 0.8006 | 0.7292 | 0.5925 | 0.6756 |
| 0.75 | 0.8073 | 0.7323 | 0.6029 | 0.684 |

- Individual mAP as weight: Each model's contribution was weighted by its mAP score.
- Uniform weights: All models were assigned an equal weight.
- Rank-based weights: Models were assigned linear or squared weights from 1 to 4 based on their performance ranking

We tested four different confidence scoring types to determine which worked best for our weapon detection system:

- 'max' (Maximum Confidence): Takes the maximum scaled confidence score in the cluster and normalizes it by the maximum model weight.
- 'avg' (Simple Average): Averages the scaled confidence scores of all boxes in the group to get the final score.
- 'box_and_model_avg' (Hybrid Average): This method rewards clusters supported by diverse models by summing unique model weights. It computes a weighted average score, then adjusts it based on the proportion of total model weight that contributed. This penalizes missed detections by high-weight [10].
- 'absent_model_aware_avg' (Penalized Average): Penalizes clusters where some models did not contribute (i.e. adds weights of absent models to the denominator). This method computes a weighted average score where the numerator is the sum of ($box\ score \times box\ weight$), and the denominator includes both the weights of This penalizes the final score if models, especially high-weight ones, miss the detection, reducing confidence proportionally to the weight of missing models [10].

From Table 2, we can draw the following observations regarding ensemble fusion and confidence scoring type:

The WBF ensemble method with confidence scoring strategy 'max' consistently outperformed individual models. Using the 'Max' confidence scoring strategy with the individual mAP as weights produced an ensemble mAP of 0.838, which is a significant improvement over the best individual performance (SSD model with VGG16 backbone). This confirms that giving more influence to higher-performing models leads to more accurate final detections.

Individual mAP weighting marginally outperformed uniform weighting (0.838 vs 0.833). This suggests that even though combining all models is beneficial for improvement over individual models, emphasizing the predictions of more accurate models leads to a marginal gain in detection performance. Rank-based weighting yielded lower mAP (0.828), indicating that overly penalizing weaker models (e.g., EfficientNet) reduces ensemble diversity.

Among the various strategies for computing the final confidence, the max method—where the strongest individual prediction dominates—delivers superior performance compared to averaging schemes, especially in challenging detection scenarios.

Table 2. Ensemble mAP for various confidence scoring types with Different Weighting Strategies: (a) model weight = mAP of model (b) Uniform weights (c) Linear Ranking: model weights ranging from 1 to 4, with the worst-performing model receiving a weight of 1 and the best-performing a weight of 4 (d) Square Ranking: model weights = rank in the increasing order of performance squared

| Confidence Scoring Types for WBF | Individual CNN based backbones | SSD model mAP for NMS Threshold = 0.55 | Weighting Strategies for mAP (@0.5 IoU) | | | |
|---|---|---|---|---|---|---|
| | | | (a) | (b) | (c) | (d) |
| max | VGG | 0.814 | 0.838 | 0.833 | 0.828 | 0.824 |
| | Resnet | 0.743 | | | | |
| | MobileNet | 0.666 | | | | |
| | EfficientNet | 0.603 | | | | |
| avg | VGG | 0.814 | 0.740 | 0.732 | 0.761 | 0.782 |
| | Resnet | 0.743 | | | | |
| | MobileNet | 0.666 | | | | |
| | EfficientNet | 0.603 | | | | |
| box_and_ model_avg | VGG | 0.814 | 0.738 | 0.729 | 0.757 | 0.777 |
| | Resnet | 0.743 | | | | |
| | MobileNet | 0.666 | | | | |
| | EfficientNet | 0.603 | | | | |
| absent_model_ aware_avg | VGG | 0.814 | 0.739 | 0.730 | 0.757 | 0.780 |
| | Resnet | 0.743 | | | | |
| | MobileNet | 0.666 | | | | |
| | EfficientNet | 0.603 | | | | |

## VI. CONCLUSION

Through this work, we present an ensemble-based weapon detection system which leverages the SSD framework. Four SSD models were independently trained using diverse CNN backbones: VGG16, ResNet50, EfficientNet, and

MobileNetV3, to capture varied spatial and semantic features of weapons. These models were trained on a curated dataset composed of three weapon classes: guns, heavy weapons (e.g., rifles, shotguns), and knives. The results of these models were then aggregated using the Weighted Boxes Fusion (WBF) method, an ensemble technique that combines bounding box proposals based on confidence-weighted overlap. The performance-weighted ensemble achieved a mean Average Precision (mAP) of 0.838, outperforming both the equal-weight ensemble (mAP: 0.833) and all standalone SSD models. Importantly, this was achieved by weighting the contribution of each model based on its individual mAP and implementing a 'max' confidence scoring strategy within the WBF. These results demonstrate the effectiveness of model diversity and adaptive weighting in enhancing detection accuracy. Potential avenues for future research include refining the ensemble architecture, exploring alternative CNN backbones within the ensemble, investigating alternative fusion methods which dynamically adjusts weights based on image conditions or incorporating attention mechanisms to guide feature extraction.